\documentclass[sigconf,authorversion]{acmart}

\usepackage{subfigure}
\usepackage{balance}

\copyrightyear{2020} 
\acmYear{2020} 
\setcopyright{acmcopyright}
\acmConference[CIKM '20] {The 29th ACM International Conference on Information and Knowledge Management}{October 19--23, 2020}{Virtual Event, Ireland}
\acmBooktitle{The 29th ACM International Conference on Information and Knowledge Management (CIKM '20), October 19--23, 2020, Virtual Event, Ireland}
\acmPrice{15.00}
\acmDOI{10.1145/3340531.3412704}
\acmISBN{978-1-4503-6859-9/20/10}

\begin{document}
\fancyhead{}

\title{Ensembled CTR Prediction via Knowledge Distillation}

\author{Jieming Zhu}
\affiliation{%
  \institution{Huawei Noah's Ark Lab}
  \city{Shenzhen}
  \country{China}
}
\email{jmzhu@ieee.org}

\author{Jinyang Liu}
\affiliation{%
\institution{School of Data and Computer Science}
  \institution{Sun Yat-Sen University, China}
}
\email{liujy57@mail2.sysu.edu.cn}

\author{Weiqi Li}
\authornote{Work done during internship at Huawei.}
\affiliation{%
\institution{School of Data and Computer Science}
  \institution{Sun Yat-Sen University, China}
}
\email{liwq23@mail2.sysu.edu.cn}

\author{Jincai Lai}
\affiliation{%
  \institution{Huawei Noah's Ark Lab}
  \city{Shenzhen}
  \country{China}
}
\email{laijincai@huawei.com}

\author{Xiuqiang He}
\affiliation{%
  \institution{Huawei Noah's Ark Lab}
  \city{Shenzhen}
  \country{China}
}
\email{hexiuqiang1@huawei.com}

\author{Liang Chen,~~~~Zibin Zheng}
\affiliation{%
\institution{School of Data and Computer Science}
  \institution{Sun Yat-Sen University, China}
}
\email{{chenliang6, zhzibin}@mail.sysu.edu.cn}

\begin{abstract}
Recently, deep learning-based models have been widely studied for click-through rate (CTR) prediction and lead to improved prediction accuracy in many industrial applications. However, current research focuses primarily on building complex network architectures to better capture sophisticated feature interactions and dynamic user behaviors. The increased model complexity may slow down online inference and hinder its adoption in real-time applications. Instead, our work targets at a new model training strategy based on knowledge distillation (KD). KD is a teacher-student learning framework to transfer knowledge learned from a teacher model to a student model. The KD strategy not only allows us to simplify the student model as a vanilla DNN model but also achieves significant accuracy improvements over the state-of-the-art teacher models. The benefits thus motivate us to further explore the use of a powerful ensemble of teachers for more accurate student model training. We also propose some novel techniques to facilitate ensembled CTR prediction, including teacher gating and early stopping by distillation loss. We conduct comprehensive experiments against 12 existing models and across three industrial datasets. Both offline and online A/B testing results show the effectiveness of our KD-based training strategy. 
\end{abstract}



\maketitle

\section{Introduction}

In many applications such as recommender systems, online advertising, and Web search, click-through rate (CTR) is a key factor in business valuation, which measures the probability that users will click (or interact with) the promoted items. CTR prediction is extremely important because, for applications with a large user base, even a small improvement of prediction accuracy can potentially lead to a large increase of the overall revenue. For example, existing studies from Google~\cite{DCN,WideDeep} and Microsoft~\cite{EnsembleBing} reveal that an absolute improvement of 1\textperthousand ~in AUC or logloss is considered as practically significant in industrial-scale CTR prediction problems. 



With the recent success of deep learning, many deep models have been proposed and gradually adopted in industry, such as Google's Wide\&Deep~\cite{WideDeep}, Huawei's DeepFM~\cite{DeepFM}, Google's DCN~\cite{DCN}, and Microsoft's xDeepFM~\cite{xDeepFM}. They have demonstrated remarkable performance gains in practice. Current research for CTR prediction has two trends. First, deep models tend to become more and more complex in network architectures, for example, by employing convolution networks~\cite{CNN-FeatureGen}, recurrent networks~\cite{DSIN}, attention mechanisms~\cite{autoint,DIN}, and graph neural networks~\cite{FiGNN} to better capture sophisticated feature interactions and dynamic user behaviors. Second, following the popular wide\&deep learning framework~\cite{WideDeep}, many deep models employ a form of ensemble of two different sub-models to improve CTR prediction. Typical examples include DeepFM~\cite{DeepFM}, DCN~\cite{DCN}, xDeepFM~\cite{xDeepFM}, AutoInt+~\cite{autoint}, etc. Although these complex model architectures and ensembles lead to improved prediction accuracy, they may slow down model inference and even hinder the adoption in real-time applications. How to make use of a powerful model ensemble to achieve the best possible level of accuracy yet still retain the model complexity for online inference is the main subject of this work.

Different from the aforementioned research work that focuses primarily on model design, in this paper, we propose a model training strategy based on knowledge distillation (KD)~\cite{KD}. KD is a teacher-student learning framework that guides the training of a student model with the knowledge distilled from a teacher model. Then, the student model is expected to achieve better accuracy than it would if trained directly. While KD has been traditionally applied for model compression~\cite{KD}, 
\textcolor{black}{we do not intentionally limit the student model size in terms of hidden layers and hidden units. We show that the training strategy not only allows to simplify the student model as a vanilla DNN model without sophisticated architectures, but also leads to significant accuracy improvements over the state-of-the-art teacher models (e.g., DeepFM~\cite{DeepFM}, DCN~\cite{DCN}, and xDeepFM~\cite{xDeepFM}).} 
Furthermore, the use of KD allows flexibility to develop teacher and student models of different architectures while only the student model is required for online serving. 

These benefits motivate us to further explore the powerful ensemble of individual models as teachers, which potentially yields a more accurate student model. Our goal is to provide a KD-based training strategy that is generally applicable to many models for ensembled CTR prediction.
Towards this goal, we have compared different KD schemes (soft label vs hint regression) and training schemes (pretrain vs co-train) for CTR prediction. 
While KD is an existing technique, we make the following extensions: 1) We propose a \textit{teacher gating} network that enables sample-wise teacher selection to learn from multiple teachers adaptively. 
2) We propose the novel use of distillation loss as the signal for \textit{early stopping}, which not only alleviates overfitting but also enhances utilization of validation data during model training. 

We evaluate our KD strategy comprehensively on three industrial-scale datasets: two open benchmarks (i.e., Criteo and Avazu), and a production dataset from Huawei's App store. The experimental results show that after training with KD, the student model not only attains better accuracy than itself but also can surpass the teacher model. 
Our ensemble distillation achieves the largest offline improvement reported so far (over 1\% AUC improvement against DeepFM and DCN on Avazu). An online A/B test shows that our KD-based model achieves an average of 6.5\% improvement in CTR and 8.2\% improvement in eCPM (w.r.t revenue) in online traffic. 



\textcolor{black}{In summary, the main contributions of this work are as follows:
\begin{itemize}
    \item Our work provides a comprehensive evaluation on the effectiveness of different KD schemes on the CTR prediction task, which serves as a guideline for potential industrial applications of KD.
    \item Our work makes the first attempt to apply KD for ensembled CTR prediction. We also propose novel extensions to improve the accuracy of the student model.
    \item We conduct both offline evaluation and online A/B testing. The results confirm the effectiveness of our KD-based training strategy.
\end{itemize}
}

In what follows, Section~\ref{sec_background} introduces the background knowledge. Section~\ref{sec:approach} describes the details of our approach. The experimental results are reported in Section~\ref{sec:exp}. We review the related work in Section~\ref{sec:relatedwork} and finally conclude the paper in Section~\ref{sec:conclusion}. 



\section{Background}\label{sec_background}
In this section, we briefly introduce the background of CTR prediction and knowledge distillation. 

\subsection{CTR Prediction}\label{sec:ctr}

The objective of CTR prediction is to predict the probability a user will click a candidate item. Formally, we define it as $\hat{y}=\sigma(\phi(x))$, where $\phi(x)$ is the model function that outputs the logit (un-normalized log probability) value given input features $x$. $\sigma(\cdot)$ is the sigmoid function to map $\hat{y}$ to $[0,1]$. 

\textcolor{black}{Deep models are powerful in capturing sophisticated high-order feature interactions to make accurate prediction. In the following, we choose four of the most representative deep models, i.e., Wide\&Deep~\cite{WideDeep}, DeepFM~\cite{DeepFM}, DCN~\cite{DCN}, and xDeepFM~\cite{xDeepFM}, since they have been successfully adopted in industry. As described in Equation~\ref{equ:WDL}$\sim$\ref{equ:xDeepFM}, all these models follow the general wide and deep learning framework, which comprises a summation of two parts: one is a deep model $\phi_{DNN}(\mathbf{x})$, and the other is a wide model. Specifically, $\phi_{LR}$ and $\phi_{FM}$ are widely-used shallow models, while $\phi_{Cross}$ and $\phi_{CIN}$ are designed to capture bit-wise and vector-wise feature interactions, respectively.} 
\begin{eqnarray}
    Wide\&Deep:~~~\phi_{WDL}({\mathbf{x}})= \phi_{LR}(\mathbf{x}) + \phi_{DNN}(\mathbf{x}) \label{equ:WDL}\\
    DeepFM:~~~\phi_{DeepFM}({\mathbf{x}})= \phi_{FM}(\mathbf{x}) + \phi_{DNN}(\mathbf{x}) \label{equ:DeepFM} \\
    DCN:~~~\phi_{DCN}({\mathbf{x}})= \phi_{Cross}(\mathbf{x}) + \phi_{DNN}(\mathbf{x}) \label{equ:DCN}\\
    xDeepFM:~~~\phi_{xDeepFM}({\mathbf{x}})= \phi_{CIN}(\mathbf{x}) + \phi_{DNN}(\mathbf{x}) \label{equ:xDeepFM} 
\end{eqnarray}

Finally, each model is trained to minimize the binary cross-entropy loss:
\begin{equation}
    \mathcal{L}_{CE}(y, \hat{y})=-\sum_{i}\big(y_i log{\hat{y_i}} + (1-y_i)log(1-\hat{y}_i)\big)~.
    \label{equ:bce}
\end{equation}


We observe that current models tend to become more and more complex in order to capture effective high-order feature interactions. For example, xDeepFM runs about 2x slower than DeepFM with the use of outer products and convolutions. Although these models use a type of ensemble of two sub-models to boost prediction accuracy, directly ensembling more models is usually too complex to apply to real-time CTR prediction problems. Our work proposes a KD-based training strategy to enable the production use of the powerful ensemble models.





\subsection{Knowledge Distillation}
Knowledge distillation (KD)~\cite{KD} is a type of teacher-student learning framework. 
The core idea of KD is to train a student model with supervision from not only the true labels but also the guidance provided by the teacher model (e.g., mimicking the output of a teacher model). The early goal of KD is for model compression~\cite{KD,Apprentice}, where a compact model with fewer parameters is trained to match the accuracy of a large model. The success of KD has led to a variety of applications in image classification~\cite{Apprentice} and machine translation~\cite{KD-translation}. Our work makes the first attempt to apply KD for ensembled CTR prediction. Instead of compressing model size only, we aim to train a unified model from different teacher models or their ensembles for more accurate CTR prediction. 





\section{Ensembled CTR Prediction}\label{sec:approach}
In this section, we describe the details of knowledge distillation for ensembled CTR prediction. 

\subsection{Overview}
\begin{figure}[!t]
  \centering
  \includegraphics[scale=0.42]{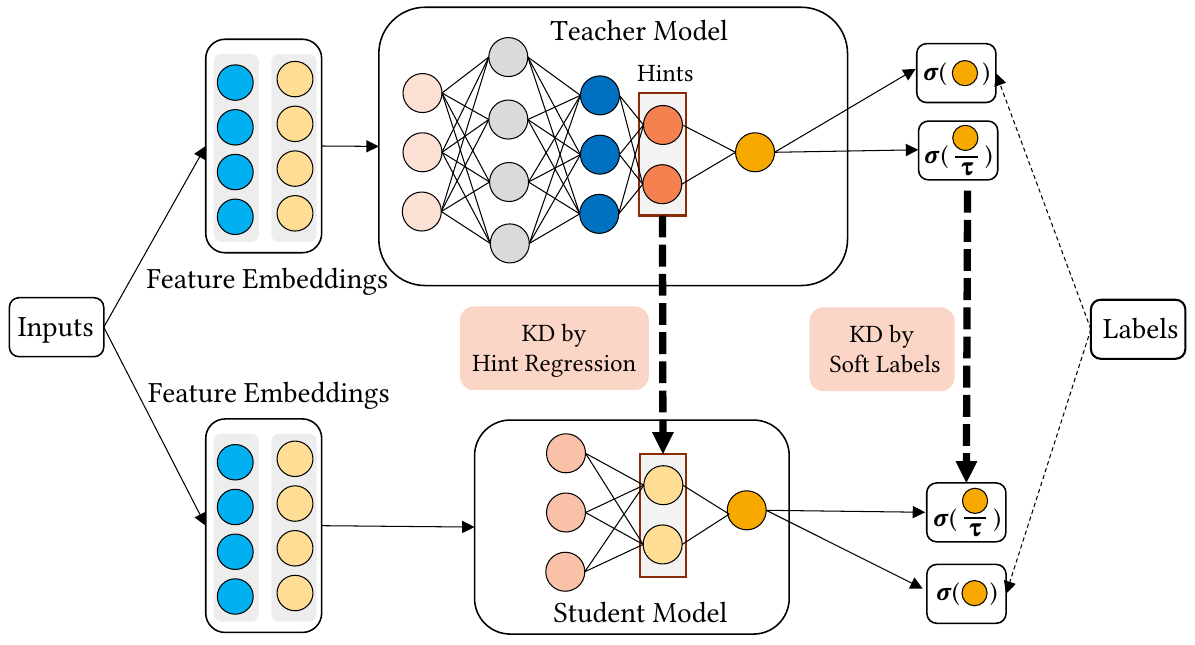}
  \caption{The framework of knowledge distillation.}\label{fig:KDF}
\end{figure}

Figure~\ref{fig:KDF} illustrates the framework of knowledge distillation, which consists of a teacher model and a student model. Let T be a teacher model that maps the input features $x$ to the click probability $\hat{y}_T$. The same features are fed to the student model $S$ and it predicts $\hat{y}_S$. During training, the loss functions for the teacher and the student models are formulated as follows:
\begin{align}
&\mathcal{L_T} = \mathcal{L}_{CE}(y, \hat{y}_T)
\label{equ:sum-teacher}\\
&\mathcal{L_S} = \gamma \mathcal{L}_{CE}(y, \hat{y}_S) + \beta\mathcal{L}_{KD}(S, T)
\label{equ:sum-student}
\end{align}
where $\mathcal{L}_{CE}(\cdot)$ denotes the binary cross-entropy loss in Equation~\ref{equ:bce} and $\mathcal{L}_{KD}(S, T)$ denotes the distillation loss (defined later in Section~\ref{sec:one_teacher}). Especially for the student loss $\mathcal{L_S}$, the first term $\mathcal{L}_{CE}$ captures the supervision from true labels and the second term $\mathcal{L}_{KD}$ measures the discrepancy between the student and the teacher models. $\beta$ and $\gamma$ are hyper-parameters to balance the weights of these two terms. Normally, we set $\beta + \gamma=1$. 

The KD framework allows flexibility to choose any teacher and student model architectures. More importantly, since only the student model is required for online inference, our KD-based training strategy does not disturb the process of model serving. In this work, we intentionally use a vanilla DNN network as the student model to demonstrate the effectiveness of KD. We intend to show that even a simple DNN model can learn well given useful guidance from a "good" teacher.

\subsection{Distillation from One Teacher}\label{sec:one_teacher}
To enable knowledge transfer from a teacher model to a student model, many different KD methods have been proposed. In this paper, we focus on the two most common methods: soft label~\cite{KD} and hint regression~\cite{FitNet}. 

\subsubsection{Soft label}\label{sec:softlabel}
In this method, the student not only matches the ground-truth labels (i.e., hard labels), but also the probability outputs of the teacher model (i.e., soft labels). In contrast to hard labels, soft labels convey the subtle difference between two samples and therefore can help the student model generalize better than directly learning from hard labels. 

As shown in Figure~\ref{fig:KDF}, the "KD by soft labels" method has the following distillation loss to penalize the discrepancy between the teacher and the student models.
\begin{equation}
\mathcal{L}_{KD}(S, T) = \mathcal{L}_{CE}\big(\sigma(\frac{z_T}{\tau}), \sigma(\frac{z_S}{\tau})\big)~,
\label{equ:softlabel}
\end{equation}
where $\mathcal{L}_{CE}$ is the binary cross-entropy loss defined in Equation~\ref{equ:bce}. $z_T$ and $z_S$ denote the logits of both models. A temperature parameter $\tau$ is further applied to producing a softer probability distribution of labels. Finally, substituting the distillation loss $\mathcal{L}_{KD}(S, T)$ in Equation~\ref{equ:sum-student} deduces the final loss for the student model, which combines the supervision of hard labels and soft labels.






\subsubsection{Hint regression}\label{sec:hintloss}
While soft labels provide direct guidance to the outputs of a student model, hint regression aims to guide the student's learning of representations. As "KD by hint regression" shown in Figure~\ref{fig:KDF}, a hint vector $\mathbf{v}_T$ is defined as an intermediate representation vector from a teacher's hidden layer. Likewise, we choose the student's representation vector $\mathbf{v}_S$ and force $\mathbf{v}_S$ to approximate $\mathbf{v}_T$ with linear regression. 
\begin{equation}
\mathcal{L}_{KD}(S, T) = \Arrowvert W\mathbf{v}_T-\mathbf{v}_S\Arrowvert^2_2~,
\label{equ:hintregression}
\end{equation}
where ${W} \in R^{n \times m}$ is a transformation matrix in case that $\mathbf{v}_T \in R^m$ and $\mathbf{v}_S \in R^n$ have different sizes. To some extent, the distillation loss $\mathcal{L}_{KD}$ acts as a form of regularization to the student model and helps the student mimic the learning process of the teacher. Accordingly, substituting the distillation loss $\mathcal{L}_{KD}(S, T)$ in Equation~\ref{equ:sum-student} derives the final loss for training the student model. 



\begin{figure}[!t]
  \centering
  \includegraphics[width=0.355\textwidth]{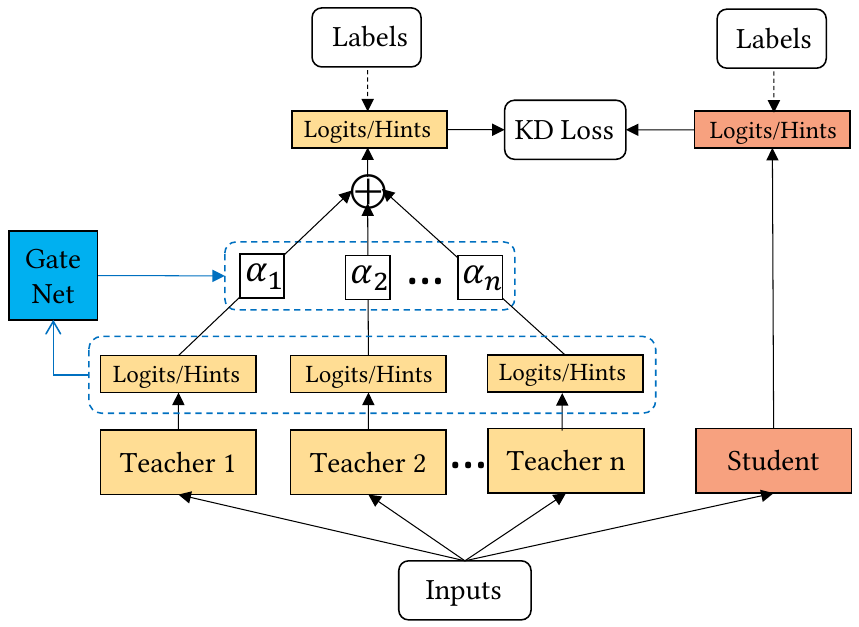}
  \caption{The framework of adaptive ensemble distillation
}\label{fig:end_framework}
\end{figure}

\subsection{Distillation from Multiple Teachers}



Model ensemble is a powerful technique to improve prediction accuracy and has become a dominant solution to win various recommendation competitions (e.g., Kaggle~\cite{KaggleEnsemble}, Netflix~\cite{NetflixPrize}). However, directly applying model ensemble is often impractical due to the high model complexity for both training and inference. To make it possible, we extend our KD framework from a single teacher to multiple teachers. 


One straightforward way is to average individual teacher models to make a stronger ensemble teacher, so that the problem reduces to learning from a single teacher. However, due to the different model architectures and training schemes used in practice, not all teachers can provide equally important knowledge on each sample. An ineffective teacher may even misguide the learning of the student. To gain effective knowledge from multiple teachers, we propose an adaptive ensemble distillation framework, as illustrated in Figure~\ref{fig:end_framework}, to dynamically adjust their contributions. Formally, we have the following adaptive distillation loss:
\begin{equation}
\mathcal{L}_{KD}(S, T) =  \mathcal{L}_{KD}(S, \sum_{i=1}^M \alpha_i T_i)~,
\label{equ:adaptive_KD_loss}
\end{equation}
where $\alpha_i$ is the importance weight for controlling the contribution of teacher $T_i$ of $M$ teachers. $\sum_{i=1}^M \alpha_i T_i$ denotes the ensemble of teachers, where the teacher weights should satisfy $\sum_{i=1}^M \alpha_i=1 $. 


\textbf{Teacher gating network}. Instead of setting $\alpha_i$ as fixed parameters, we intend to learn $\alpha_i$ dynamically and make it adaptive to different data samples. To achieve this, we propose a teacher gating network to adjust the importance weights of teachers, which enables sample-wise teacher selection. More specifically, we employ a softmax function as the gating function.
\begin{equation}
\alpha_i = \frac{exp(w_iz_{T_i}+b_i)}{\sum_{i=1}^Mexp(w_iz_{T_i}+b_i)}~,~~~~ i=1,\cdots,M
\end{equation}
where $\{w_i, b_i\}_{i=1}^M$ are the parameters to learn. In intuition, the gating network uses all teachers' outputs to determine the relative importance of each other. 

\subsection{Training}\label{sec:training}

\subsubsection{Training scheme}
\textcolor{black}{In this work, we investigate both the pre-train scheme and the co-train scheme for KD. The pre-train scheme is to train the teacher and student models in two phases. For the co-train scheme, both teacher and student models are trained jointly while the back-propagation of the distillation loss is unidirectional so that the student model learns from the teacher, but not vice versa. The co-train scheme is faster than the two-phase pre-train scheme, but requires much more GPU memory. We also empirically compare the performance between pre-train and co-train schemes in Section~\ref{sec:RQ1}. 
For the case of ensemble distillation, we focus mainly on the pre-train scheme, since the co-training scheme requires to load multiple teacher models into the GPU memory. We consider three state-of-the-art models (i.e., DeepFM, DCN, xDeepFM) as teachers and train each of them by minimizing the loss in Equation~\ref{equ:sum-teacher}. When training the student model according to Equation~\ref{equ:sum-student}, the teacher can provide better guidance to the student model. The teacher gating network is trained together with the student model for sample-wise teacher selection. }

\subsubsection{Early stopping via distillation loss}\label{sec:early_stop}
It is a common practice to employ early stopping to reduce overfitting. When training a teacher model, we employ the left-out validation set to monitor the metrics (e.g., AUC) for early stopping. When training a student model, we instead propose to use the distillation loss from the teacher model as the signal for early stopping. It works because the teacher model has been trained with reduced overfitting by early stopping on the validation set, and the student model can inherit this ability when mimicking the teacher model. This eliminates the need to retain the validation set and helps the student model generalize better by using the validation set (often the most recent data) to enrich the training data.





\section{Experiments}\label{sec:exp}
In this section, we report on our experimental results to evaluate the effectiveness of our KD-based training strategy for CTR prediction.

\subsection{Experiment Setup}
\subsubsection{Datasets} 
We use three real-world datasets in our experiments: Criteo\footnote{https://www.kaggle.com/c/criteo-display-ad-challenge}, Avazu\footnote{https://www.kaggle.com/c/avazu-ctr-prediction}, and our production dataset. 
Table \ref{tab:Data Statis} summarizes the data statistics information.

\textbf{Criteo}. The dataset consists of ad click logs over a week. It comprises 26 categorical feature fields and 13 numerical feature fields. Following Google's DCN work~\cite{DCN}, we randomly split the data of the last two days into a validation set and a test set of equal size, while the rest is used for training.

\textbf{Avazu}. The dataset contains 10 days of click logs. It has a total of 22 fields with categorical features such as app id, app category, device id, etc. 
Following the recent AutoInt work~\cite{autoint}, we randomly split the data into 8:1:1 as the training, validation, and test sets, respectively.

\textbf{Production}. We collected the production dataset from users' click logs over 8 days. It has 29 categorical feature fields, such as app id, category, tags, city, recent clicks, etc. Following our practice in production, we split the data sequentially, where the first 7 days are used for training and the last day is equally split for validation and testing, respectively.

For the two open datasets, we preprocess the data following the work in~\cite{autoint}. First, we replace the features less than a threshold with a default ``<UNK>'' ID, where the threshold is set to 10 and 5 for Criteo and Avazu, respectively. Second, numerical values in Criteo are normalized to avoid large variance by transforming each value $x$ to $\log^2(x)$, if $x > 2$.

\begin{table}[!t]
\small
\centering
\caption{Dataset statistics}\label{tab:Data Statis} \vspace{-1ex}
\begin{tabular}{c||c|c|c|c}
\hline
{Dataset}& \#Instances & \#Fields & \#Features & Positive Ratio \\
\hline 
 Criteo  & 46M & 39 & 1M & 26\% \\
\hline
  Avazu & 40M & 22 & 972K & 17\% \\
\hline
  Production & 231M & 29 & 160K & -- \\
\hline
\end{tabular}
\vspace{-3ex}
\end{table}





\subsubsection{Baseline models} 
To compare the performance, we select a total of 12 representative models, including LR, FM~\cite{FM}, FFM~\cite{FFM}, DNN~\cite{DNN}, Wide\&Deep~\cite{WideDeep}, DeepFM~\cite{DeepFM}, DCN~\cite{DCN}, xDeepFM~\cite{DeepFM}, PIN~\cite{PIN}, FiBiNet~\cite{FiBiNET}, AutoInt+~\cite{autoint}, and FGCNN~\cite{CNN-FeatureGen}. Although we cannot enumerate all the existing models, we have made a relatively comprehensive comparison in contrast to previous studies. To test statistical significance, in Section~\ref{sec:RQ4}, we repeat the experiments 5 times by changing random seeds, and run the two-tailed paired t-test.

\subsubsection{Implementation details} 
All the models are implemented in PyTorch. We use Adam for optimization, and set the batch size to 2000. The learning rate is set to 0.001. Categorical features are embedded to dimensions of 20, 40, and 40 for Criteo, Avazu, and Production, respectively. We use grid search to tune other hype-parameters. Specifically, we apply $L_2$ regularization on feature embeddings and search the regularization weight in $[0, 1e-8, 1e-7, 1e-6, 1e-5]$. We also apply dropout to hidden layers of DNN with dropout rates selected in $[0.1, 0.2, 0.3, 0.4, 0.5]$. The number of hidden layers is set among $[2, 3, 4, 5, 6]$. We keep the same size of each layer and select it from $[300, 400, 500, 600]$. Especially, we vary the layers of the cross net (in DCN) and CIN (in xDeepFM) among $[1,2,3,4,5]$. As for KD by soft labels, we set $\beta + \gamma=1$ and tune $\beta$ from $0$ to $1$ with a step of $0.1$. For KD by hint regression, we set $\gamma=1$ and tune $\beta$ in the range $[0, 1e-6, 1e-5, 1e-4, 1e-3]$. We also try different values of temperature $\tau$ among $[1,2,3,4,5,7,10,15]$. To avoid overfitting, early stopping is adopted when the metrics on the validation set (for teacher training) or the distillation loss (for student training) stop improving in three consecutive epochs. 

\begin{table}[!t]
  \centering
  \caption{Performance of different KD schemes. T$\rightarrow$S indicates that T is a teacher and S is a student.}\vspace{-1ex}
  \label{sec4-2}
  \includegraphics[width=0.48\textwidth]{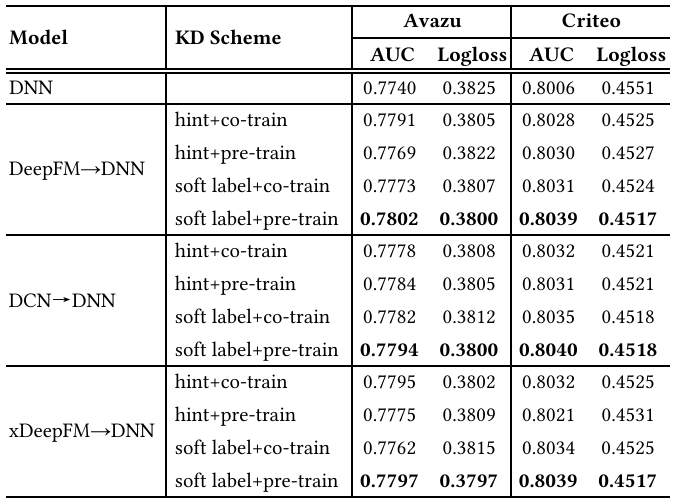}
  \vspace{-2ex}
\end{table}
\begin{table*}[!t]
  \centering
  \caption{Performance across different teacher and student models. Each column represents a teacher model (T) and each row represents a student model (S). 3T denotes an ensemble of teachers: DeepFM + DCN + xDeepFM. The row "w/o KD" represents the performance of using teachers only. LL is short for logloss. The best results in each column are marked in bold.}
  \label{sec4-3}
  \includegraphics[width=1\textwidth]{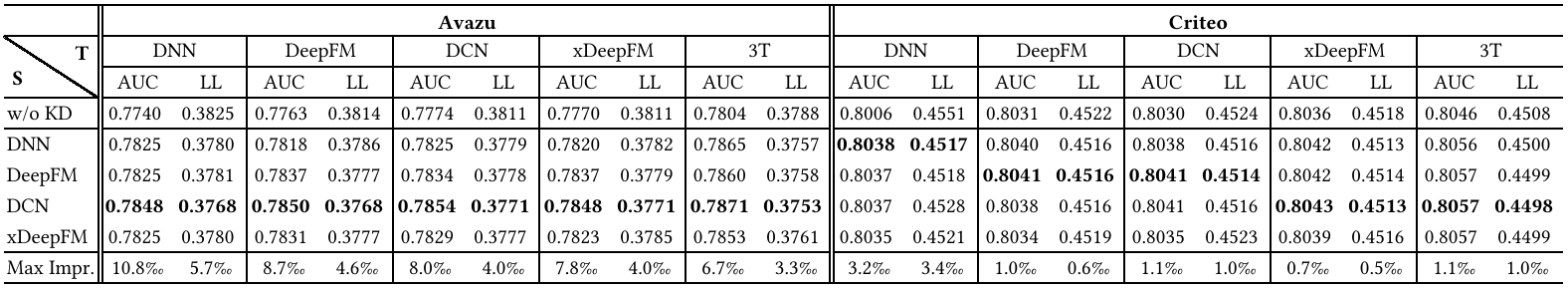}
\end{table*}
\subsection{Performance of Different KD Schemes}\label{sec:RQ1}

In this section, we aim to study the research question of which KD scheme performs the best for CTR prediction. Here, we define a candidate KD scheme as a combination of the KD methods (soft label v.s. hint regression) and training methods (pre-train v.s. co-train). To study the effect of KD schemes only, in this experiment, we fix the size of the student DNN model as $500\times5$. Each model is tuned to attain its best performance for Avazu and Criteo, respectively. 



Table~\ref{sec4-2} shows the results of different KD schemes. Especially, the first row indicates the student-only performance, i.e., training the DNN directly. We can make the following observations from the results: 
1) All student models trained via different KD schemes consistently outperform the student-only DNN by a large margin. 
2) The KD scheme of "soft label + pre-train" performs the best among all candidate KD schemes. However, the difference between these schemes is small, making it a flexible choice to select an appropriate KD scheme. 
In practice, the co-train scheme requires to load both of the teacher model and the student model into the GPU memory, which hinders its scalability especially when multiple teacher models are available. Therefore, we focus mainly on the "soft label + pre-train" scheme in the following experiments.



\subsection{Performance across Different Teachers and Students}\label{sec:RQ2}
In this experiment, we intend to evaluate what performance could be achieved by applying different teacher models and student models. In particular, we select five teacher models including DNN, DeepFM, DCN, xDeepFM, and 3T (an ensemble of DeepFM + DCN + xDeepFM). We first tune each teacher model to attain its best performance, and report the teachers' results as the row "w/o KD" in Table~\ref{sec4-3}. The results of 3T are an average of the three models. Then, we explore different student models (i.e., DNN, DeepFM, DCN, xDeepFM) and perform KD experiments in a total of 20 settings (5 teachers and 4 students). We fix the teacher models and tune the student models. 

The experimental results are reported in row 2$\sim$5 in Table~\ref{sec4-3}. We observe that: 1) All student models trained with KD achieve even better performance than those using teachers only. The last row shows the maximal improvements (absolute value in \textperthousand~) over the teacher's performance. This finding (i.e., students beating teachers) is suprising since it is rarely reported in traditional KD studies, where a student model is usually cut to a small size for model compression. We further evaluate the impact of student model size on performance in Section~\ref{sec:RQ6-1}. 2) With our KD strategy, even the vanilla DNN model can learn well, surpassing the state-of-the-art complex teacher models. This confirms the good model capacity of DNN and the effectiveness of KD to guide the learning. Using DeepFM or DCN as a student can even attain better performance. This shows that better student architectures can be further explored to maximize the benefit of KD. But we focus mainly on the DNN model to demonstrate the superiority of KD, and leave the design of an optimal student model architecture for future research. 3) Compared to the cases of KD from one teacher, the performance is largely improved by using the 3T ensemble as teacher models. In total, 3T$\rightarrow$DNN makes up to $12.5$\textperthousand~and $5$\textperthousand~absolute improvements in AUC over a single teacher model for Avazu and Criteo, respectively. The encouraging results motivate us to explore more towards ensemble distillation.


\subsection{Performance of Different Teacher Ensembles}\label{sec:RQ3}
After confirming the superiority of using 3T as an ensemble of teachers, we intend to evaluate the detailed performance of using different numbers and different types of teacher ensembles. As shown in Table~\ref{sec4-4}, we experiment with different number of teachers from 1T to 6T, and report both the results of teacher ensembles and their students. For the case 1T, the results correspond to using DCN and xDeepFM as the single teacher for Avazu and Criteo respectively, due to their best performance. For the cases of multiple teachers (3T and 6T), we study two types of training methods (denoted as M and D) to generate teachers. Especially, 3T(M) denotes the combination "DeepFM + DCN + xDeepFM" with different model architectures. 6T(M) doubles 3T(M) with different random initialization seeds. 3T(D) and 6T(D) denote the combination of 3 or 6 models (DCN for Avazu and xDeepFM for Criteo) trained on different data partitions of training and validation sets, while the testing set remains the same. Again, we observe that student models trained with different teacher ensembles consistently outperform their teachers. Meanwhile, applying more teachers (1T$\rightarrow$3T$\rightarrow$6T) for ensemble distillation results in higher performance of student models. But the increase diminishes. In addition, teacher ensembles trained on different data partitions perform better (i.e., D better than M), which in turn leads to better students. This is partly due to the reason that we use the best-performing single teachers for 3T(D) and 6T(D) settings.

\begin{table}[!t]
  \centering
  \caption{Performance of different teacher ensembles. $x$T indicates $x$ teachers. M and D denote two types of ensemble model training, via different model architectures (M) or different data partitions (D). The best results in each setting is marked in bold. }\vspace{-1ex}
  \label{sec4-4}
  \includegraphics[width=0.4\textwidth]{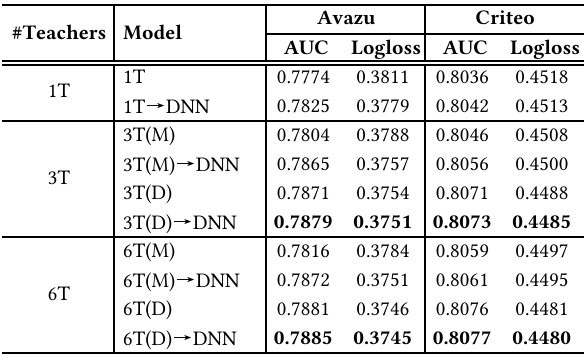}
  \vspace{-1ex}
\end{table}
\begin{table}[!t]
  \caption{Performance comparison between our DNN models and the state-of-the-art models (The upper part shows our experimental results and the lower part reports the results from existing work).}\vspace{-1ex}
  \label{tab: compare_sota}
  \begin{flushleft}{The best baseline is underlined while our best result is marked in bold.~~\textbf{*} indicates $p<0.001$ by two tailed paired t-test with 5 runs.} \end{flushleft}
  \includegraphics[width=0.38\textwidth]{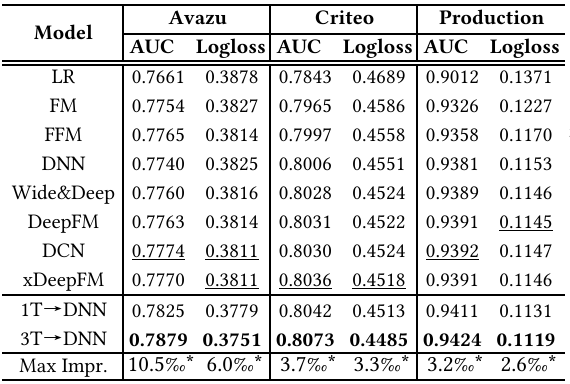}
  
  \vspace{1ex}
  \begin{flushleft}{A comparison of some recent deep models with respect to the performance improvements over DeepFM (absolute values in \textperthousand~).} \end{flushleft}
   \includegraphics[width=0.31\textwidth]{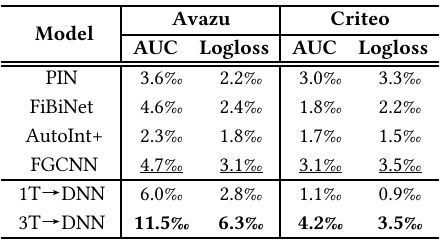}
   \vspace{-3ex}
\end{table}

\subsection{Comparison with the State-of-the-art Models}\label{sec:RQ4}
We make a comprehensive performance comparison between our DNN models trained with KD and the state-of-the-art models (12 models in total). The upper part of Table~\ref{tab: compare_sota} shows the overall performance of all the compared models across three datasets. In particular, the underlined numbers show the best baseline models and bold numbers are the best results among all the models. 1T and 3T in the table denote the best teacher models mentioned in Table~\ref{sec4-4}. With a focus on evaluating KD on representative models, we did not reproduce all of the baseline models. Instead, we compare some recent deep models directly using the results reported in their papers, which are shown in the lower part of Table~\ref{tab: compare_sota}. For fairness, we only show the performance improvements over DeepFM. 

We summarise our observations from the table as follows:  
1) Deep models generally outperform shallow models (LR, FM, and FFM), which reveals the capability of deep models to more capture complex feature interactions.  2) Models trained with KD (1T and 3T) consistently beat all baseline models with a large margin on all the three datasets. Specifically, 3T makes $10.5$\textperthousand~, $3.7$\textperthousand~, and $3.2$\textperthousand~absolute improvements in AUC over the best-performing baseline models for the three datasets. 
3) 3T makes further improvements compared to 1T on all datasets. This confirms the value of ensemble distillation, which makes the student models generalize better given the diversity of different teachers.   
4) Compared to some more recently proposed deep models as shown in the lower part, the largest improvements from 3T also confirms the effectiveness of KD, considering that our student model is a vanilla DNN. Finally, we emphasize that a student model trained with KD can even surpass the ensembled teacher models is a surprising finding. It deserves our more investigation on the KD mechanism to help reduce overfitting in future.

\subsection{Ablation Study}\label{sec:RQ5}
\subsubsection{Teacher gating}
We evaluate the effectiveness of teacher gating by comparing the performance of training with gating or without gating (i.e., via averaging), in the case of 3T and 6T on Avazu. To demonstrate the capability of gating to distill effective knowledge from diverse teachers, in this experiment, we set 6T as an ensemble of DeepFM, xDeepFM, DCN, DNN, Wide\&Deep and AutoInt+, and 3T as DeepFM, xDeepFM, and DCN. From the results shown in Figure~\ref{fig:ablation_study}(a), we see that the student model learns better with teacher gating, which indicates the usefulness of sample-wise teacher selection. It is worth noting that, without gating, 6T performs worse than 3T because 6T contains more diverse models. Teacher gating helps adjust the importance weights of teachers adaptively and thus makes 6T largely improved.

\subsubsection{Early stopping via distillation loss}
\begin{figure}[t!]
\centering    
\subfigure[Teacher gating] 
{
    \centering          
    \includegraphics[scale=0.48]{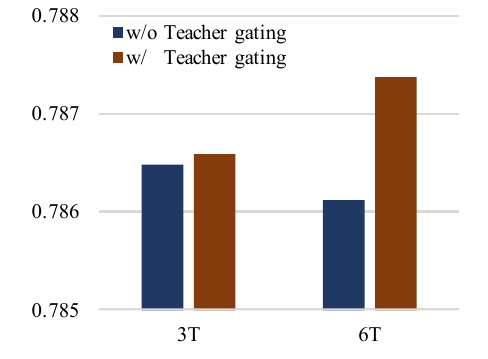}  
}
\subfigure[Early stop via distillation loss] 
{
    \centering      
    \includegraphics[scale=0.48]{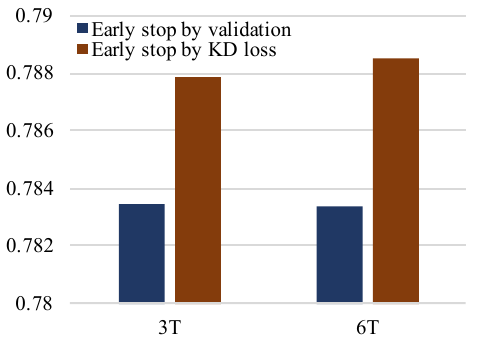}
}
\vspace{-1ex}
\caption{Ablation study results in AUC} 
\label{fig:ablation_study}  
\end{figure}

We compare the two ways of early stopping (using validation set v.s. using KD loss). The results in Figure~\ref{fig:ablation_study}(b) show that the KD loss between the teacher and student models serves a good monitor signal for early stopping. It allows full utilization of the validation data (usually more recent) for training the student model and thus obtains better performance. 


\begin{table}[!t]
  \centering
  \caption{Performance comparison of different student sizes. The upper part describes the model sizes and the lower part shows the corresponding results.}
  \label{sec4-7-1}
   \includegraphics[width=0.45\textwidth]{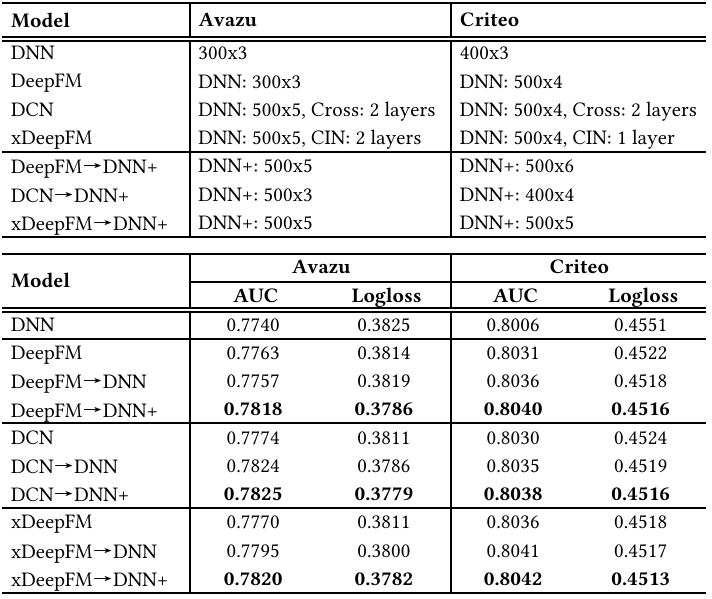}
\end{table}

\subsection{Impact of student model size}\label{sec:RQ6-1}
The size of a DNN model, w.r.t. both hidden layers and hidden units, usually determines its model capacity. To study the impact of student model size, we first tune teacher models and keep them fixed (the same settings with Section~\ref{sec:RQ2}). We then perform two group of experiments. In the first group, we set each student model to its best DNN size tuned in the last step (denoted as DNN). In the second group, we re-tune the size of each student DNN model to a larger size (denoted as DNN+). Table~\ref{sec4-7-1} presents the detailed model sizes in the upper part and the the corresponding results in the lower part. For instance, the DNN model alone attains the best performance at the size of 300x3 (3 hidden layers with 300 units each) on Avazu. However, when using DeepFM as a teacher for KD training, the 500x5 DNN+ performs better than the above 300x3 DNN. We can observe similar results in other settings from the table. 
The results imply that our goal is not directly model compression, since a larger student model tends to achieve better performance.

\subsection{A/B Testing}\label{sec:ckd}
We conducted an online A/B test in the app recommender system of Huawei's App store. It has hundreds of millions of daily active users which generates hundreds of billions of user feedback events everyday, such as browsing, clicking and downloading apps. In the online serving system, hundreds of candidate apps are first selected during the matching phrase, and then ranked by a ranking model (e.g., DCN) to generate the final recommendation list. 
In the A/B test, the baseline model in production is an extended version of DCN, demonstrating superiority over other competing models. As the model is re-trained on a daily basis, we choose the recent two old versions of models as ensemble teachers for efficiency consideration and then distill a student model of the same architecture on the newest data over recent 15 days for online serving. The A/B test was performed over one week, and we randomly select 5\% of online traffic for both the control group and our experiment group. On average, our model improves the overall DLR (app downloading rate defined as $\#download/\#impressions$) by 6.5\% and eCPM  (expected cost per mille) by 8.2\% over the baseline. This is a substantial improvement and demonstrates the effectiveness of our KD approach. 

\section{Related Work}\label{sec:relatedwork}
\subsection{CTR Prediction}



Since the successful application of DNNs to Youtube recommendation~\cite{DNN}, deep learning has been widely studied for CTR prediction. Some work investigates the integration of DNNs and traditional shallow models (e.g., Wide\&Deep~\cite{WideDeep}, DeepFM~\cite{DeepFM}). Some models aim to capture different orders of feature interactions explicitly (e.g., DCN~\cite{DCN}, xDeepFM~\cite{xDeepFM}). Some models explore the use of convolutional networks (e.g., FGCNN~\cite{CNN-FeatureGen}), recurrent networks (e.g., DSIN~\cite{DSIN}), attention networks (e.g., AutoInt~\cite{autoint}, FiBiNET~\cite{FiBiNET}) and graph neural networks (e.g., FiGNN~\cite{FiGNN}) to learn high-order feature interactions. Some other studies focus on modeling evolutionary user interests in CTR prediction (e.g., DIN~\cite{DIN}). Our work demonstrates the effectiveness of KD using multiple representative models, but the framework is generally applicable to other models since they can be used as teacher models as well.



Although model ensemble is a powerful technique to boost prediction accuracy, the high complexity in such ensembles (e.g., over 100 models ensembled in the Netflix prize~\cite{NetflixPrize}) hinders the adoption in the industry. To restrict the model complexity for online deployment, most industrial studies reported by Facebook~\cite{GBDTLR}, Google~\cite{GoogleEnsemble}, and Microsoft~\cite{EnsembleBing,DeepGBM} focus on the ensemble of only two models. 
Instead, our work tackles this issue via ensemble distillation and delivers a simplified yet equally powerful student model for online inference. 

\subsection{Knowledge Distillation}
The knowledge distilled from a teacher varies in different forms. In addition to soft label~\cite{KD} and hint regression~\cite{FitNet} we used, some work further explores the knowledge transfer through inter-layer flow~\cite{KDGift}, cross-sample similarity~\cite{CrossSampleKD}, attention mechanism~\cite{attention}, etc. Some more recent efforts have been made towards ensemble distillation, facilitating a variety of tasks such as image classification~\cite{MEAL} and machine translation~\cite{KD-translation}.
For recommender systems, Chen et al.~\cite{ChenZXQZ19} propose an adversarial distillation approach to learn from external knowledge base for collaborative filtering. Xu et al.~\cite{PrivilegedKD} proposes a KD approach to transfer privileged features from ranking to matching tasks. Liu et al.~\cite{KD_debias} leverage KD for debiasing in recommendation via uniform data. Zhang et al.~\cite{Distill_wsdm} study the mutual learning between path-based and embedding based recommendation models. 
Our work is partially inspired from these studies and serves as the first attempt to apply KD for ensembled CTR prediction.

\section{Conclusion}\label{sec:conclusion}
Model ensemble is a powerful technique to improve prediction accuracy. In this paper, we make an attempt to apply KD for ensembled CTR prediction. Our KD-based training strategy enables the production use of a powerful ensemble of teacher models and makes large accuracy improvements. Surprisingly, we demonstrate that a vanilla DNN trained with KD can even surpasses the ensemble teacher models. 
\textcolor{black}{Our key contribution includes an intensive evaluation of the KD-based training strategy, teacher gating for sample-wise teacher selection, and early stopping by KD loss that increases the utilization of validation data. }
Both offline and online A/B testing results demonstrate the effectiveness of our approach. We hope that our encouraging results could attract more research efforts to study training strategies for CTR prediction.

\section{Acknowledgements}
The work described in this paper was partially supported by the Key-Area Research and Development Program of Guangdong Province (2018B010109001), and the National Natural Science Foundation of China (U1811462).

\bibliographystyle{ACM-Reference-Format}
\bibliography{cikm2020}
\balance
\end{document}